\documentclass[10pt, a4paper]{article}
\usepackage{lrec}
\usepackage{multibib}
\newcites{languageresource}{Language Resources}
\usepackage{graphicx}
\usepackage{tabularx}
\usepackage{soul}
\usepackage{epstopdf}
\usepackage[latin1]{inputenc}

\usepackage{hyperref}
\usepackage{xstring}

\usepackage{slashbox}
\usepackage{color}

\usepackage{enumitem}
\usepackage{multirow}
\usepackage{hhline}
\usepackage{graphicx}
\usepackage{caption}
\usepackage{subcaption}
\usepackage{amsmath}

\newtheorem{theorem}{Theorem}
\newtheorem{definition}[theorem]{Definition}

\newcounter{example}%[section]
\newenvironment{example}[1][]{\refstepcounter{example}\par\medskip
   \noindent {\theexample. #1} \rmfamily}{\medskip}
   
\title{\textbf{Structured Interpretation of Temporal Relations}}

\name{Yuchen Zhang and Nianwen Xue}
\address{Brandeis University \\
         415 South Street, Waltham, MA \\
         yuchenz@brandeis.edu, xuen@brandeis.edu\\}
\abstract{
Temporal relations between events and time expressions in a document are often modeled in an unstructured manner 
where relations between individual pairs of time expressions and events are considered in isolation. This often results in inconsistent and incomplete annotation and computational modeling. We propose a novel annotation approach where events and time expressions in a document form a dependency tree in which each dependency relation corresponds to an instance of temporal anaphora where the antecedent is the parent and the anaphor is the child.
We annotate a corpus of 235 documents using this approach in the two genres of news and narratives, with 48 documents doubly annotated. We report a stable and high inter-annotator agreement on the doubly annotated subset, validating our approach, and perform a quantitative comparison between the two genres of the entire corpus. We make this corpus publicly available. %\footnote{https://github.com/yuchenz/structured\_temporal\_relations\_corpus} %We plan to make this data set publicly available after it has gone through a rigorous validation process.
 \\ \newline  }

\begin{document}

\maketitleabstract

\section{Introduction}

Understanding temporal relations between events and temporal expressions in a natural language text is a fundamental part of understanding the meaning of text. Automatic detection of temporal relations also enhances downstream natural language applications such as story timeline construction, question answering, text summarization, information extraction, and others.  
%Temporal relation modeling is one of the fundamental tasks in information extraction and text understanding. By recognizing the temporal relations between events and time expressions in a document, it not only provides a more complete machine understanding of the text, but also helps downstream applications such as semantic analysis, story timeline construction, question answering, entailment resolution, summarization, etc. 
Due to its potential, temporal relation detection has received a significant amount of interest in the NLP community in recent years. 

Most of the research attention has been devoted to defining the ``semantic'' aspect of this problem -- the identification of a set of semantic relations between pairs of events, between an event and a time expression, or between pairs of time expressions. Representative work in this vein includes TimeML \cite{pustejovsky2003timeml}, a rich temporal relation markup language that is based on and extends Allen's Interval Algebra \cite{allen1984towards}. TimeML has been further enriched and extended for annotation in other domains  \cite{o2016richer,styler2014temporal,mostafazadeh2016caters}. Corpora annotated with these schemes \cite{pustejovsky2003timebank,o2016richer} are shown to have stable inter-annotator agreements, validating the temporal relations proposed in the TimeML. Through a series of TempEval shared tasks \cite{tempeval2007,tempeval2010,tempeval2012,tempeval2015,tempeval2016,tempeval2017}, there has also been significant amount of research on building automatic systems aimed at predicting temporal relations. 

Less attention, however, has been given to the ``structural'' aspect of temporal relation modeling -- answering the question of which other events or time expressions a given time expression or event depends on for the interpretation of its temporal location. Having an answer to this question is important to both linguistic annotation and computational modeling. From the point of view of linguistic annotation, without an answer to this question, an annotator is faced with the choice of: (i) labeling the relation between this event/time expression with all other events and time expressions, or (ii) choosing another event/time expression with which the event/time expression in question has the most salient temporal relation. (i) is impractical for any textual document that is longer than a small number of sentences. Without  a solid linguistic foundation, adopting (ii) could lead to inconsistent and incomplete annotation as annotators may not agree on which temporal relations are the most salient. 

From a computational perspective, without knowing which time expressions and events are related to each other, an automatic system has to make a similar choice to predict the temporal relations between either all pairs of events and time expressions, or only a subset of the temporal relations. If it chooses to do the former, there will be ${n}\choose{2}$ pairs for $n$ events and time expressions. Not only is this computationally expensive, there could be conflicting predictions due to the transitivity of temporal relations (e.g. ``A before B" and ``B before C" imply ``A before C", which a pair-wise approach may make conflicting predictions) and additional steps are necessary to resolve such conflicts \cite{chambers2008jointly,yoshikawa2009jointly,do2012joint}. 
%\textcolor{red}{TO DO: If it chooses to do the latter... ???}

%Most of these work follow TimeML's scheme and model temporal relations among events and time expressions in a pair-wise fashion. Thusly, given a document with all events and time expressions marked out, each pair of them is considered and a specific temporal relation label (e.g. before, after, etc.) is assigned to the pair. %Due to temporal relations' transitivity (``A before B" and ``B before C" can imply ``A before C"), in most previous work, only a small set of crucial relations is manually annotated, and labels on the remaining pairs are inferred by computing the temporal closure automatically. This process greatly alleviates the number of annotation points and improves efficiency and consistency. Inter-annotator agreements are then computed on the temporal closured results. 
%However, we observed that there are %modeling the temporal relation between {\em every} pair of events and time expressions 
%both linguistic and computational challenges with this approach.

We propose a novel annotation approach to address this dilemma. Specifically we propose to build a dependency tree structure for the entire document where the nodes of the tree are events and time expressions, as well as a few pre-defined ``meta'' nodes that are not anchored to a span of text in the document %(for more on pre-defined nodes, see \S\ref{pre-nodes}). 
The building blocks of this dependency structure are pairs of events and time expressions in which the {\it child}  event/time expression depends on its {\it parent} event/time expression for its temporal interpretation. The dependency relation is based  on the well-established notion of {temporal anaphora} where an event or time expression can only be interpreted with respect to its {reference time}  \cite{reichenbach47,partee1973some,partes1984nominal,hinrichs1986temporal,webber1988tense,bohnemeyer2009temporal}. In each dependency relation in our dependency structure, the parent is the {\it antecedent} and the child is the {\it anaphor} that depends on its antecedent for its temporal interpretation.
Consider the following examples: 

\begin{example}
\label{timex} 
He arrived on \underline{Thursday}. He got here at \underline{8:00am}. 
\end{example}
\vspace*{-2mm}
\begin{example}
\label{event} 
He \underline{arrived} at school, \underline{walked} to his classroom, and then the class \underline{began}.
\end{example}

In (\ref{timex}), the antecedent is ``Thursday'' while ``8:00am'' is the anaphor. We won't know when exactly he arrived  unless we know the 8:00am is on Thursday. In this sense, ``8:00am'' depends on ``Thursday'' for its temporal interpretation. We define the {\em antecedent of an event as a time expression or event with reference to which the temporal location of the anaphor event can be most precisely determined.}
In (\ref{event}), the antecedent for the event ``the class began'' is ``walked to his classroom'' in the sense that the most specific temporal location for the event ``the class began'' is after he walked to the classroom. Although ``the class began'' is also after ``he arrived at school'', the temporal location we can determine based on that is not as precise.
%how about an example where the antecedent of an event if a time expression

% From a linguistic perspective, human interpretation of the temporal location of an event is intuitively not based on its temporal relations with all the other events and time expressions in the document. On the contrary, it's usually based on its temporal relation with {\em only one} other event or time expression (usually previously mentioned), to which it is most saliently related. In other words, the temporal anaphora of the current event makes reference to its temporal antecedent. For example, in sentence (a), readers would naturally locate the ``class began" event after the ``walked to his classroom'' event, instead of after the ``arrived at school'' event.
% \begin{enumerate}[label=(\alph*)]
% \item[(a)] He {arrived at school}, { walked to his classroom}, and then {the class began}.
% \end{enumerate}
% The idea of temporal anaphora is a well-established concept in formal and computational semantics \cite{partee1973some,partes1984nominal,hinrichs1986temporal,webber1988tense,bohnemeyer2009temporal}. However, no modeling schemes or annotation experiments based on this line of theory have been done.

In order for the events and time expressions to form a dependency tree, one key assumption we make is {\it there is exactly one antecedent event/time expression for each anaphor}. This ensures that there is exactly one head for each dependent, a key formal condition for a dependency tree. 

% Using a dependency structure to represent temporal relations in a document has been proposed before \cite{kolomiyets2012extracting}, but our work is more comprehensive and linguistically grounded in the following ways. First, they only work with events, not including time expressions. Time expressions are a strong source of temporal location information for events and excluding them will result in an incomplete temporal structure. We cover both events and time expressions to form a complete temporal structure for a piece of text. Second, they exclude some stative events such as modalized events, while we provide a more complete temporal structure including stative events. Third, although they link events in a text to form a dependency structure, but do not explicitly spell out the linguistic basis for the temporal dependencies and annotators are only instructed to identify the most plausible parent for each event. In contrast, we explicitly specify how antecedents of events or time expressions are determined based on a long line of theoretical and computational linguistic research \cite{reichenbach47,partee1973some,partes1984nominal,hinrichs1986temporal,webber1988tense,bohnemeyer2009temporal,Wuyun201638} and these specifications are given to annotators are given to annotators as guidelines when they annotated the data.  And lastly, their annotation work is only performed on children's stories (narrative data), while our annotated corpus covers both news and narrative genres.

Once this dependency structure is acquired, manually or automatically, additional temporal relations may be inferred based on the transitive property of temporal relations, but we argue that this dependency structure is an intuitive starting point that makes annotation as well as the computational modeling more constrained and tractable.

%To handle these challenges better, we propose a new annotation scheme which interprets temporal relations in a document as a dependency tree structure. We utilize the temporal anaphora idea and determine only one most salient reference time for each event or time expression, better representing the human intuitions on temporal relation understanding and reasoning. Specifically for time expressions, we propose a new structure capturing their temporal relations in our scheme. Instead of labeling before, overlap, etc. relations between time expressions as in most previous work, we pick the time expression that the current time expression's normalization most closely depends on as its reference time. 
%We claim that this approach is more intuitive and straightforward, and easier for downstream tasks such as story timeline construction. More details describing this new design is in Section~\ref{timex-ref}.

%Furthermore, the proposed dependency tree structure models $n$ relations for $n$ events and time expressions, which inherently prevents the conflicting prediction problem, and is less expensive computationally both for automatic systems and downstream applications. A detailed description of our annotation scheme is in Section~\ref{scheme}. 

We annotate a corpus of 235 documents with temporal dependency structures, with 48 documents double-annotated to evaluate inter-annotator agreement. The annotated data are chosen from two different genres, new data from the Xinhua newswire portion of the Chinese TreeBank \cite{xue2005nle} and Wikipedia news data used for CoNLL Shared Task on Shallow Discourse Parsing in 2016 \cite{xue2016conll}, and narrative story data from Grimm fairy tales. The two genres are chosen because the temporal structure of texts from those two genres unfolds in very different ways: news reports are primarily in report discourse mode in the sense of \cite{smith2003modes}  while Grimm fairy tales are primarily in narrative mode and time advances in those two genres in very different ways, as we will discuss in more detail in Section \ref{cross-genre}.
We report a stable and high inter-annotator agreement for both genres, which validates the intuitiveness of our approach. This corpus is publicly available.\footnote{https://github.com/yuchenz/structured\_temporal\_relations\_corpus} %after we have done a final round of proof-reading and validation.
%the temporal anaphora intuition. 
%We further annotated 235 documents with this scheme, providing a corpus available for automatic systems training and evaluation. 
%Another contribution of this work is that, while most previous work focus on data in one domain and haven't offered insights on how temporal relations differ across domains, we selected data from two different domains (narratives and news reports), and compared and analyzed their temporal differences. 
%With event types annotated, we also analyzed how different event types behave in different patterns temporally. 

The main contributions of this paper are:
\begin{itemize}
\item We propose a novel and comprehensive temporal dependency structure to capture temporal relations in text.
\item We analyze different types of time expressions in depth and propose a novel definition, as far as we know, for the reference time of a time expression (\S\ref{timex-ref}).
\item We produce an annotate corpus with this temporal structure that covers two very different genres, news and narratives and achieved high inter-annotator agreements for each genre. An analysis of the annotated data show that temporal structures are very genre-dependent, a conclusion that has implications for how the temporal structure of a text can be parsed. 
\end{itemize}
In the next few sections, we will briefly discuss related work (\S\ref{related}), describe our annotation scheme (\S\ref{scheme}), and present our annotation experiments (\S\ref{anno-sec}).
We summarize our work in \S\ref{conclusion}

\section{Related Work}
\label{related}
Using a dependency structure to represent temporal relations in a document has been proposed before \cite{kolomiyets2012extracting}, but our work is more comprehensive and linguistically grounded in the following ways. First, their dependency structure is based on events, to the exclusion of time expressions. Time expressions are a strong source of temporal location information for events and excluding them will result in incomplete temporal structures. We cover both events and time expressions to form a complete temporal structure for a text. Second, they exclude  stative events such as modalized events, while we provide a more complete temporal structure that include stative events. Third, although they link events in a text to form a dependency structure, they do not explicitly spell out the linguistic basis for the temporal dependencies and annotators are only instructed to identify the most plausible parent for each event. In contrast, we explicitly specify how antecedents of events or time expressions are determined based on a long line of theoretical and computational linguistic research \cite{reichenbach47,partee1973some,partes1984nominal,hinrichs1986temporal,webber1988tense,bohnemeyer2009temporal,Wuyun201638} and these specifications are given to annotators as guidelines when they annotated the data.  And lastly, their annotation work is only performed on children's stories (narrative data), while our annotated corpus covers both news and narrative genres. Annotating two different genres is crucial for us to show that the temporal structure for the two genres are very different, an observation that has implication for automatic parsing strategies.

% Introduction (Background, our work description, contribution, paper structure)
% why we want to do it in a structured fashion
% strong motivation
% novel idea, contribution
% old way: pair, not structured

% final goal: timeline --> need a tree --> need to find a reference time for each event/timex

% assumption: every timex, every event has one reference time; what can be a reference time? -- (some types of timex -- temporal locations; some types of events -- events, NOT states, NOT duration --> previous work is not correct)

% previous work doesn't use reference time concept, they just do pair-wise, need further resolution to do a timeline

% not all pairs are related, one node is related to only one reference time, not related to all other nodes

% every document's temporal relations have a structure, and it's based on the assumption

% structure based on one reference time

\section{Temporal Structure Annotation Scheme}
\label{scheme}
In our annotation scheme, a temporal dependency tree structure is defined as a 4-tuple $(T, E, N, L)$, where $T$ is a set of time expressions, $E$ is a set of events, and $N$ is a set of pre-defined ``meta'' nodes not anchored to a span of text in the document. $T$, $E$, $N$ form the nodes in the dependency structure, and $L$ is the set of edges in the tree. Detailed descriptions for each set are in the following subsections, followed by some examples.

\subsection{Nodes in the temporal dependency tree}
The nodes in a temporal dependency tree includes time expressions, events, and a set of pre-defined nodes. We elaborate on each type of nodes below:

\subsubsection{Time Expressions}

%Most previous work annotate all time expressions and the temporal relations related to them. However, we argue that not all time expressions contribute to the temporal structure of a document. 
TimeML \cite{pustejovsky2003timeml} treats all temporal expressions as markable units and classifies them into three categories: fully specified temporal expressions (``June 11, 1989'', ``Summer, 2002''); underspecified temporal expressions (``Monday'', ``next month'', ``last year'', ``two days ago''); and durations (``three months'', ``two years''). 
The purpose of our dependency structure annotation is to find all time expressions that can serve as a reference time for other events or time expressions. 
We observe that while the first two TimeML categories of time expressions can serve as reference times, the last category, ``durations", typically don't serve as reference times, unless they are modified by expressions like ``ago'' or ``later''. For example, the ``10 minutes'' in (\ref{modified-duration}) can serve as a reference time because it can be located in a timeline as a duration from 8:00 to 8:10, while the ``10 minutes'' in (\ref{duration}) can't serve as a reference time.

\begin{example}
\label{modified-duration}
He arrived at 8:00am. \underline{10 minutes later}, the class began.
\end{example}
\vspace*{-6mm}
\begin{example}
\label{duration} 
It usually takes him \underline{10 minutes} to bike to school.
\end{example}

Therefore, in our annotation scheme, we make the distinction between time expressions that can be used as reference times and the ones that cannot. The former includes fully specified temporal expressions, underspecified temporal expressions, as well as time durations modified by ``later'' or ``ago''. The latter include unmodified durations. In  our annotation, only the former are considered to be valid nodes in our time expression set $T$.

\subsubsection{Events}

We adopt a broad definition of events following \newcite{pustejovsky2003timeml}, where ``an event is any situation (including a process or a state) that happens, occurs, or holds to be true or false during some time point (punctual) or time interval (durative).'' Based on this definition, unless stated explicitly, events for us include both eventive and stative situations. Adopting the minimal span approach along the lines of  \cite{o2016richer}, only the headword of an event is labeled in actual annotation. Since different events tend to have different temporal behaviors in how they relate to other events or time expressions\cite{Wuyun201638}, we also assign a coarse event classification label to each event before linking them to other other events or time expressions to form a dependency structure. Adapting the inventory of situation entity types from \newcite{smith2003modes} and  from \newcite{zhang2014automatic}, we define the following eight categories for events.

\begin{itemize}
 \item An {\bf Event} is a process that happens or occurs. It is the only eventive type in this classification set that advances the  time in a text. An example event is ``I {\em went} to school yesterday''.
 \item A {\bf State} is a situation that holds during some time  interval. It is stative and describes some property or state of an object, a situation, or the world. For example, ``she {\em was} very shy'' describes a state. 
 \end{itemize}
 The remaining event types are all statives that describe  an eventive process. 
 \begin{itemize}
 \item A {\bf Habitual} event describes the state of a regularly repeating event, as in ``I {\em go} to the gym three times a week''. 
 \item An {\bf Ongoing} event describes an event in progress, as in ``she was {\em walking} by right then''. 
 \item A {\bf Completed} event describes the completed state of an event, as in  ``She's {\em finished} her talk already''. 
 \item A {\bf Modalized} event describes the capability, possibility, or necessity of an event, as in ``I have to {\em go}''. 
 \item A {\bf Generic Habitual} event is a Habitual event for generic subjects, as in ``The earth {\em goes} around the sun''.
\item A {\bf Generic State} is a state that hold for a generic subject, as in ``Naked mole rats don't {\em have} hairs''. 
 
All valid events from a document, represented by their headwords, form the event set $E$.

\end{itemize}
\subsubsection{Pre-defined Meta Nodes}
\label{pre-nodes}
In order to provide valid reference times for all events and time expressions, and to form a complete tree structure, we designate the following pre-defined nodes for the set $N$.

{\bf ROOT} is the root node of the temporal dependency tree and every document has one ROOT node. It is the parent of (i) all other pre-defined nodes, and (ii) absolute concrete time expressions (Example \ref{root-parent}, see \S\ref{timex-ref} for more on time expression classification). The meta node {\bf DCT} is the Document Creation Time, a.k.a. Speech Time. Following \newcite{pustejovsky2003timeml}, we define meta nodes {\bf PRESENT\_REF, PAST\_REF, FUTURE\_REF} as the general reference times respectively for generic present, past, and future times. Lastly, {\bf ATEMPORAL} is designated as the parent node for atemporal events, such as timeless generic statements (Example \ref{atemporal-parent}).

%provide examples of when these generic reference times are needed as antecedents and explain why DCT is not sufficient for that purpose.

These generic reference times are necessary for time expressions and events that don't have a more specific reference time in the text as their parents. For example, it is common to start a narrative story with a few descriptive statements in past tense without a specific time (Example \ref{past-ref-event}), or a general time expression referring to the past (Example \ref{past-ref-timex}). Both cases take ``Past\_Ref'' as their parent. 
\begin{example}
\label{past-ref-event}
It \underline{was} a snowy night. [Past\_Ref]
\end{example}
\vspace*{-3.5mm}
\begin{example}
\label{past-ref-timex}
\underline{Once upon the time}, ... [Past\_Ref]
\end{example}

It is worth noting that ``DCT'' and ``Present\_Ref'' are not interchangeable. ``DCT'' is usually a very specific time-stamp such as ``2018-02-15:00:00:00'', while ``Present\_Ref'' is a general temporal location reference.  We use ``DCT'' as the parent for relative concrete time expressions (example \ref{dct-parent}), and for vague time expressions, their antecedent is ``Present\_Ref'' (Example \ref{present-ref-parent}). See \S\ref{timex-ref} for more details on time expression classification.
\begin{example}
\label{present-ref-parent}
China annual economic output results have grown increasingly smooth in \underline{recent years}. [Present\_Ref]
\end{example}
\vspace*{-3mm}
\begin{example}
Economists who try to estimate actual growth \underline{tend} to come up with lower numbers. [Present\_Ref]
\end{example}
\vspace*{-3mm}
\begin{example}
China will remain a trade partner as important to Japan as the United States \underline{in the future}. [Future\_Ref]
\end{example}
\vspace*{-3mm}
\begin{example}
\label{dct-parent}
The economy expanded 6.9 percent \underline{last year}. [DCT]
\end{example}
\vspace*{-3mm}
\begin{example}
\label{root-parent}
A trend of gradual growth began in \underline{2011}. [ROOT]
\end{example}
\vspace*{-3mm}
\begin{example}
\label{atemporal-parent}
The earth \underline{goes} around the sun. [Atemporal]
\end{example}

\subsection{Edges in the temporal dependency tree}

As we discussed above, each dependency relation consists of an antecedent and an anaphor, with the antecedent being the parent and the anaphor being the child. Based on the well-established notion of temporal anaphora \cite{reichenbach47,partee1973some,partes1984nominal,hinrichs1986temporal,webber1988tense,bohnemeyer2009temporal}, we assume each event or time expression in the dependency tree has only one antecedent (i.e. one reference time), which is necessary to form the dependency tree.
In this section, we will first discuss  what can serve as a reference time for time expressions in our annotation scheme, then we will discuss what can be a reference time for events. All links between events/time expressions and their reference times form our link set $L$.
% Although an event or a time expressions might have temporal relations with multiple other events or temporal locations around them, we argue that intuitively only one previously mentioned event or temporal location is the reference time that readers use for the interpretation of the current event's temporal location. We propose to build a complete temporal dependency structure by linking each event and temporal location to their reference times in the document. The set of all such links form the set $L$.

\begin{table*}
\centering
\begin{tabular}{c|c|c|c|c|c}
\hline
\multicolumn{4}{c|}{\bf Taxonomy} & {\bf Examples} & {\bf Possible Reference Times} \\\hline
 & Locatable & \multirow{2}{*}{Concrete} & Absolute & May 2015 & ROOT \\\cline{4-6} 
Time & Time  & & Relative & today, two days later & DCT, another Concrete \\  \cline{3-6}
 Expressions & Expressions &  \multicolumn{2}{c|}{Vague} & nowadays & Present/Past/Future\_Ref \\\cline{2-6}
& \multicolumn{3}{c|}{Unlocatable Time Expressions} & every month & - \\ \hline
\end{tabular}
\caption{\label{timex-taxonomy} Taxonomy of time expressions in our annotation scheme, with examples and possible reference times.}
\end{table*}

\subsubsection{Reference Times for Time Expressions}
\label{timex-ref}

In previous work such as the TimeBank \cite{pustejovsky2003timeml} the temporal relations between time expressions are annotated with temporal ordering relations such as ``before'', ``after'', or ``overlap'' just like events in a pair-wise without considering the dependencies between them. For example, consider the three time expressions ``2003'', ``March'', and ``next year'' in (\ref{relative-absolute}), using a pair-wise annotation approach, three temporal relations will be extracted:
%\begin{enumerate}[label=(\alph*)]
%\item[(e)] 
\begin{example}
\label{relative-absolute}
The economy expanded 6.6 percent in \underline{2003}$_{t1}$, reaching its peak 7.1 percent in \underline{March}$_{t2}$. The growth rate doubled in the \underline{next year}$_{t3}$.
\end{example}
%\end{enumerate}
\begin{quote}
(2003, includes, March) \\
(2003, before, next year) \\
(March, before, next year)
\end{quote}

We argue the sole purpose for annotating temporal relations between time expressions is to properly ``interpret'' time expressions that ``depend'' on another time expression for their interpretation. In the context of time expressions, ``interpretation''  means normalizing time expressions in a format that allows the ordering between the time expressions to be automatically computed. Time expression normalization is necessary in many applications. For example, in a question answering system, our model needs to be able to answer ``2004'' when it is asked ``Which year did China's export rate double?'', instead of answering ``next year'' which is uninterpretable taken out of the original context. In order for the time expressions to be properly interpreted, it is important to annotate the dependency between ``March'' and its reference time ``2004'' because the former depends on the latter for its interpretation. Similarly, it is also important to establish the dependency between ``next year'' and its reference time ``2004'' as we won't know which year is ``next year'' until we know it is with reference to ``2004''.  With the these dependencies identified and the time expressions normalized, the temporal  relations between all pairs of  time expressions in a text can be automatically computed, and explicit annotation of the temporal relation between all pairs of time expressions  will not be necessary. For example, with ``March'' normalized to ``2003-03'' and ``next year'' normalized to ``2004'', the relation between 2003-03 and 2004 can be automatically computed. We argue that this notion of reference time for time  expressions is intuitive and easy to define. Annotating temporal dependency between time expressions is also more efficient than annotating the temporal ordering between all pairs of time expressions.

%We observe that (i) this representation of temporal relations includes redundant information. ``2003 includes March'' and ``2003 before next year'' include the information ``March before next year'' and redundant annotation means unnecessary annotation effort for human annotators.

Based on these considerations, we propose a novel definition of the reference time for time expressions:
\begin{definition}
Time expression A is the reference time for time expression B, if B depends on A for its temporal location interpretation.
\end{definition}
In other words, a time expression can depend solely on its reference time to be interpreted and normalized. We use a generic {\bf Depend-on} label for these relations. Take (\ref{timex}) as an example, annotators only need to determine that the temporal  interpretation of `8am'' depends on ``Thursday''. With ``Thursday'' normalized to, for example, ``2003-04-05'', we can then compute a normalized time ``2003-04-05:08:00:00'' for ``8am'', and easily compute the temporal ordering between them: (``2003-04-05'' includes ``2003-04-05:08:00:00'').

% With this design of reference times for time expressions, annotators have a simpler and clearer annotation goal, and temporal relations between time expressions are represented more concisely without redundant information, and more comprehensively. It also naturally supports further time expression normalization very well.
% However, on one hand, we argue that a full timeline with all temporal locations normalized to a unified format is a clearer and more comprehensive way of representing the ordering and relations among temporal locations; on the other hand, we aim at building a tree structure which has better computational properties. Therefore, we propose to not capture before, overlap, etc. relations between temporal locations, and define the reference time for a temporal location to be the closest previously mentioned temporal location that the normalization for the current temporal location depends on. These links are annotated with a {\bf Depend-on} label. 

We now consider the question of what types of nodes can serve as the reference time or antecedent for a time expression. First, since a time expression relies on its reference time for its temporal interpretation, naturally an event cannot serve as its reference time. Second, since some time expressions (e.g., ``2003'') can be interpreted (and normalized) on its own without any additional information, while others can not, further categorization of time expressions is needed to precisely specify  which time expressions need a reference time for their interpretation and which do not,  and what  time expressions can serve as reference times and which do not.

First, we make the distinction between Concrete and Vague time expressions. A {\bf Concrete Time Expression} is a time expression that can be located onto a timeline as an exact time point or interval, e.g. ``June 11, 1989'', ``today''. Their starting and ending temporal boundaries on the timeline can be determined. A {\bf Vague Time Expression} (e.g., ``nowadays'', ``recent years'', ``once upon the time'') expresses the concept of (or a period in) general past, general present, or general future, without specific temporal location boundaries. The reference time for Vague time expressions are the pre-defined nodes PRESENT\_REF, PAST\_REF, and FUTURE\_REF. 

Concrete time expressions are further classified into {\bf Absolute Time Expressions} and {\bf Relative Time Expressions}, corresponding to fully-specified (``June 11, 1989'', ``Summer, 2002'') and underspecified temporal expressions (``Monday'', ``Next month'', ``Last year'', ``Two days ago'') in \newcite{pustejovsky2003timeml} respectively. Relative concrete time expressions take either DCT or another concrete time expression as their reference time. Absolute concrete time expressions can be normalized independently and don't need a reference time. Therefore, we stipulate that their parent in the dependent tree is the pre-defined node ROOT. For example, ``1995'', ``20th century'' are absolute concrete time expressions, while ``today'', ``last year'', ``the future three years'', ``January 20th'', ``next Wednesday'' are relative concrete time expressions, and ``recent years'', ``in the past a few years'', ``nowadays'', ``once upon the time'' are vague time expressions.

An example of a concrete relative time expression having a concrete absolute temporal expression as its reference time is given in (\ref{relative-absolute}) . Consider the time expression ``March''. In order to be able to interpret it and normalize it into a valid temporal location on a timeline, we need to establish  ``2003" is its reference time. Then it is possible to normalize it into a formal representation  as ``2003-03''.
%\begin{enumerate}[label=(\alph*)]
%\item[(e)] 
% \begin{example}
% \label{relative-absolute}
% In \underline{2003}$_{t1}$, China's GDP increased by 6.6\%. The growth rate reached its peak 7.1\% in \underline{March}$_{t2}$, and the export rate doubled in the \underline{next year}$_{t3}$.
% \end{example}
%\end{enumerate}

Lastly, in order to form a complete tree structure, all pre-defined nodes (except for ROOT) take ROOT as their parent. A complete taxonomy of time expressions in our annotation scheme with examples and their possible reference times is illustrated in Table~\ref{timex-taxonomy}.
%and an example of a fully linked tree structure for temporal locations is shown in part of Figure~\ref{parse-example}.
%Under this scheme, the local temporal dependency tree structure for the three temporal expressions in  \ref{relative-absolute} is as shown in Figure~\ref{timex-graph-b}. Since $t1$, $t2$, and $t3$ are all absolute concrete temporal locations, they are linked to ROOT directly \footnote{Here and after, link labels between temporal locations and pre-defined nodes (i.e. ``Depend-on'' and ``Root'' lables) are omitted in our examples.}.

%\begin{figure}[h]
%\centering
%\includegraphics[scale=0.5]{timex-taxonomy.png}
%\caption{Taxonomy of Temporal Expressions in Our Annotation Scheme. Possible reference time nodes are listed in boxes next to each type of temporal location.}
%\label{}
%\end{figure}

\subsubsection{Reference Times for Events}
\label{event-ref}

The reference time for an event is a time expression or pre-defined node or another event with respect to which the most specific temporal location of the event in question can be determined. 
Unlike time expressions, for which the possible reference times can only be other time expressions or pre-defined nodes, the possible reference times for events are not as restrictive and can be any of the three categories. The dependency relation that we use to characterize the relationship  between the reference time / antecedent and an event is a temporal relation between them. 

\begin{definition}
Time expression/pre-defined node/event A is the reference time for event B, if A is the most specific temporal location which B depends on for its own temporal location interpretation.
\end{definition}

There has been significant amount of work attempting to characterize the temporal relationship between events, and between time expressions and events.
One of the first attempts to model temporal relations is Allen's Interval Algebra theory \cite{allen1984towards}. They introduced a set of distinct and exhaustive temporal relations that can hold between two time intervals, which are further adapted and extended in \newcite{pustejovsky2003timeml}, THYME \cite{styler2014temporal}, etc. A detailed comparison of these sets can be found in \newcite{mostafazadeh2016caters}. Mindful of the need to produce consistent annotation, and in line with the practice of some prior work such as the TempEval evaluations \cite{verhagen2007semeval,verhagen09,verhagen2010semeval}
we adopt a simplified set of  4 temporal relations to characterize the relationship between an event and its reference time. The set of temporal relations we use with their mappings to their corresponding TimeML temporal relations are shown shown in Table~\ref{timeml_set}.

\begin{table}[!h]
\begin{center}
\begin{tabular}{lll}
\hline
\bf Our Scheme & & \bf TimeML   \\ \hline
Before & & Before, IBefore \\ \hline
After & &  - \\ \hline
Overlap	& & Ends, Begins, Identity, Simultaneous \\\hline
Includes & & During \\\hline
\end{tabular}
\end{center}
\caption{\label{timeml_set} Our temporal relation set for events with mappings to TimeML's set. }
\end{table}

Although an event can in principle take a time expression, another event, or a pre-defined node as its antecedent, different types of events have different tendencies as to the types of antecedents they take.
An eventive event usually takes either a time expression or another eventive event as its reference time. They advance the time  in the narrative of a text, so it usually has a (time expression, Includes, event) relation with its antecedent, or a (event, Before, event) relation. For example, in (\ref{timex}) the time expression ``Thursday'' has  ``Includes'' relation with the event ``arrived'', and the time expression ``8:00am'' has an ``Includes'' relation with the event ``got here''. And in (\ref{event}) the event ``arrived'' has a  ``Before'' relation with the event ``walked''.

A stative event can take a time expression, another event, or a pre-defined node (except for ROOT) as its reference time. It generally describes a state that holds during the time indicated by its antecedent time expression, event, or generic time.  It usually has an ``Overlap'' relation with their reference times. For example, in (\ref{duration}) the event ``takes'' is a stative Habitual event, which describes a state of the present situation for ``him'', so its reference time is the pre-defined node ``Present\_Ref'', and has an ``Overlaps'' relation with ``Present\_Ref''.

An eventive event rarely takes a stative event as its reference time. As discussed above, we pick the most specific temporal location as the reference time for an event. Since more specific temporal locations are usually available (such as another eventive event), a stative event rarely serves as the reference time for an eventive event.

Readers are referred to our more detailed guidelines\footnote{https://github.com/yuchenz/structured\_temporal\_relations\_corpus} on time expression and event recognition, classification, and reference time annotation, which details basic principles for specific cases and discusses extra rules for special scenarios.

\subsection{Full Temporal Structure Examples}

\begin{figure}[h]
%\begin{centering}
\centering
\includegraphics[scale=0.13]{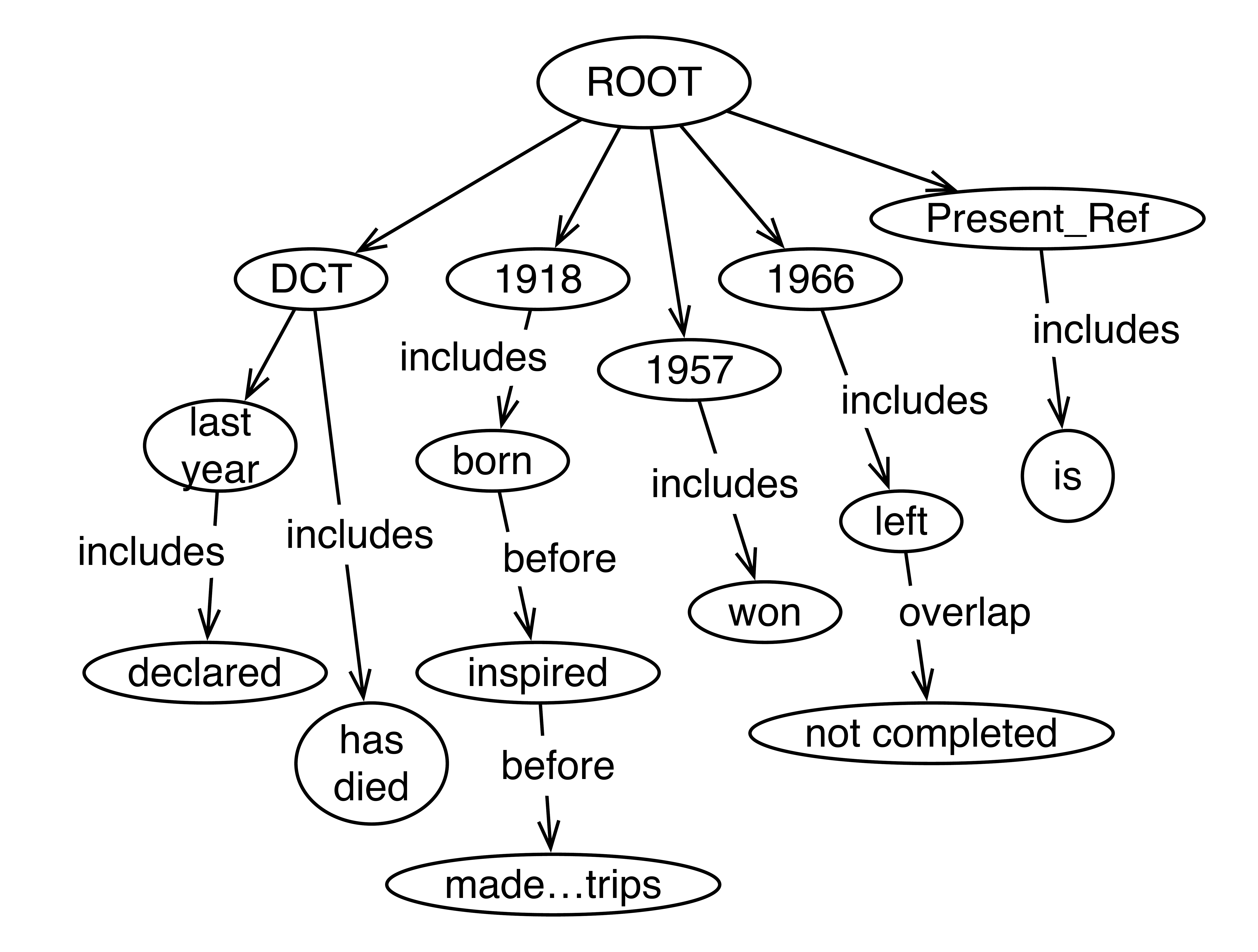}
\caption{An example full temporal dependency structure for news paragraph (\ref{utzon}).}
\label{full-parse-news}
%\end{centering}
\end{figure}

\begin{figure}[h]
%\begin{centering}
\centering
\includegraphics[scale=0.6]{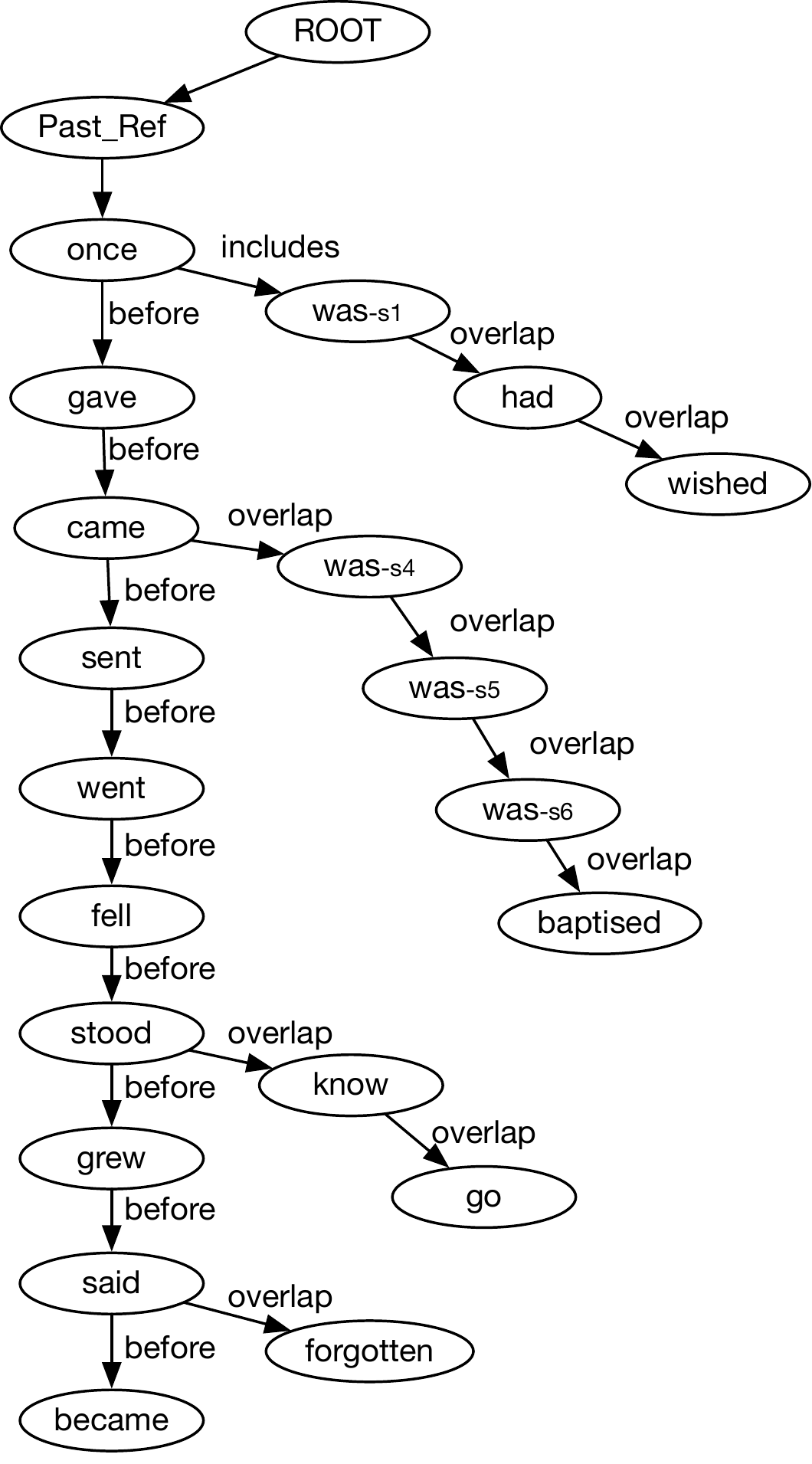}
\caption{An example full temporal dependency structure for narrative paragraph (\ref{grimm-example}).}
\label{full-parse-grimm}
%\end{centering}
\end{figure}

We present a full example temporal dependency structure for a short news report paragraph (\ref{utzon}), as illustrated in Figure~\ref{full-parse-news}, and another one for a narrative passage (\ref{grimm-example}), as illustrated in Figure~\ref{full-parse-grimm}. Subscript $e$ denotes eventive events, $t$ denotes time expressions, and $s$ denotes stative events. Unlabeled edges are ``depend-on'' relations.

The two examples provide a sharp contrast between the typical temporal dependency structures for newswire documents and narrative stories, with the former generally having a flat and shallow structure and the latter having a narrow and deep structure.

%\begin{enumerate}[label=(\alph*)]
\begin{example}
\label{utzon}
Jorn Utzon, the Danish architect who designed the Sydney Opera House, has \underline{died}$_{e1}$ in Copenhagen. \underline{Born}$_{e2}$ in \underline{1918}$_{t1}$, Mr Utzon was \underline{inspired}$_{e3}$ by Scandinavian functionalism in architecture, but \underline{made a number of inspirational trips}$_{e4}$, including to Mexico and Morocco. In \underline{1957}$_{t2}$, Mr Utzon's now-iconic shell-like design for the Opera House unexpectedly \underline{won}$_{e5}$ a state government competition for the site on Bennelong Point on Sydney Harbour. However, he \underline{left}$_{e6}$ the project in \underline{1966}$_{t3}$. His plans for the interior of the building \underline{were not completed}$_{s1}$. The Sydney Opera House \underline{is}$_{s2}$ one of the world's most classic modern buildings and a landmark Australian structure. It was \underline{declared}$_{e7}$ a UNESCO World Heritage site \underline{last year}$_{t4}$. \footnote{From a news report on {\em The Telegraph}}
\end{example}
%\end{enumerate}

\begin{example}
\label{grimm-example}
There \underline{was}$_{s1}$ \underline{once}$_{t1}$ a man who had seven sons, and still he \underline{had}$_{s2}$ no daughter, however much he \underline{wished}$_{s3}$ for one.  At length his wife again \underline{gave}$_{e1}$ him hope of a child, and when it \underline{came}$_{e2}$ into the world it \underline{was}$_{s4}$ a girl.  The joy \underline{was}$_{s5}$ great, but the child  \underline{was}$_{s6}$ sickly and small, and had to be privately \underline{baptized}$_{s7}$ on account of its weakness.  The father \underline{sent}$_{e4}$ one of the boys in haste to the spring to fetch water for the baptism.  The other six \underline{went}$_{e5}$ with him, and as each of them wanted to be first to fill it, the jug \underline{fell}$_{e6}$ into the well.  There they \underline{stood}$_{s8}$ and did not \underline{know}$_{s9}$ what to do, and none of them dared to \underline{go}$_{s10}$ home.  As they still did not return, the father \underline{grew}$_{e7}$ impatient, and \underline{said}$_{e8}$, they have certainly \underline{forgotten}$_{s11}$ it while playing some game, the wicked boys. He \underline{became}$_{e9}$ afraid that the girl would have to die without being baptized.\footnote{From Grimm's fairy tale {\em The Seven Ravens}}
%, and in his anger cried, I wish the boys were all turned into ravens.  Hardly was the word spoken before he heard a whirring of wings over his head, looked up and saw seven coal-black ravens flying away.
\end{example}

%\begin{table*}
%\centering
%\begin{tabular}{c|rrr|rrr}
%\hline
% & \multicolumn{3}{c|}{News} & \multicolumn{3}{c}{Narratives} \\ \cline{2-7}
% & Single & Double & Total & Single & Double & Total \\ \hline
%\# Docs &  91 & 24 & & 96 & 24 \\
%\# Sentences & 2,271 & 570 & & 2,903 & 797 \\
% \# Tokens & 45,132 & 11,132 & & 77,299 & 17,456 \\
%\# Timex & 901 & 265 & & 92 & 40 \\ 
%\# Events & 3,758 & 1,047 & & 8,362 & 1,952  \\ \hline
%\end{tabular}
%\caption{\label{data-stat} Corpus annotation statistics. ({\em Timex} stands for time expressions.) }
%\end{table*}

\subsection{Annotation Process}

We use a two-pass annotation process for this project. In the first pass, annotators do temporal expression recognition and classification, and then reference time resolution for all time expressions. The purpose of this pass is to mark out all possible reference times realized by time expressions and recognize their internal temporal relations, in order to provide a backbone structure for the final dependency tree. In the second pass, event recognition and classification, and then reference time resolutions for all events are annotated, completing the final temporal dependency structure of the entire document.

\begin{table*}
\centering
\begin{tabular}{cc|c|c|c|c}
\hline
%\backslashbox{\bf child}{\bf parent} 
& & Pre-defined Node & Time Expression  & Eventive Event & Stative Event \\\hline
& Time Expression & 1078 (92\%) & 89 (8\%)& 0 & 0 \\
News & Eventive Event & 103 (9\%) &  290 (26\%) & 716 (65\%) & 0 \\
& Stative Event & 149 (8\%) & 192 (11\%) & 432 (24\%) & 1029 (57\%) \\\hline
& Time Expression & 95 (83\%) & 20 (17\%) & 0 & 0 \\
Narratives & Eventive Event & 20 (0\%)  &   25 (1\%) & 4875 (99\%) & 0 \\
& Stative Event & 25 (1\%) & 74 (2\%) & 1655 (49\%) & 1612 (48\%) \\
\hline
\end{tabular}
\caption{\label{dist-parent-child} Distribution of parent types for each child type. Rows represent child types, and columns represent parent types.}
\end{table*}

\section{Annotation Analysis}
\label{anno-sec}
\subsection{Corpus}

A corpus of 115 news articles, sampled from Chinese TempEval2 data \cite{tempeval2010} and Wikinews data, \footnote{zh.wikinews.org} and 120 story articles, sampled from Chinese Grimm fairy tales, \footnote{https://www.grimmstories.com/zh/grimm\_tonghua/index} are compiled and annotated. 20\% of the documents are double annotated by native Chinese speakers. Table~\ref{data-stat} presents the detailed statistics. High and stable inter-annotator agreements are reported in Table~\ref{IAA}. 

%\begin{table*}
%\centering
%\begin{tabular}{cc|rrrrr}
%\hline
% & & \# Docs & \# Sentences & \# Tokens & \# Timex & \# Events \\ \hline
% & Single & 91 & 2,271 & 45,132 & 901 & 3,758 \\
%News & Double & 24 & 570 & 11,132 & 265 & 1,047 \\
%& Total & 115 & 2,841 & 56,264 & 1,166 & 4,805\\ \hline
%& Single & 96 & 2,903 & 77,299 & 92 & 8,362 \\
%Narratives & Double & 24 & 797 & 17,456 & 40 & 1,952 \\ 
%& Total  & 120 & 3,700 & 94,755 & 132 & 10,314 \\ \hline
%\end{tabular}
%\caption{\label{data-stat} Corpus annotation statistics. ({\em Timex} stands for time expressions.) }
%\end{table*}

\begin{table}[h]
\centering
\begin{tabular}{cc|rrrr}
\hline
 & & Docs & Sent  & Timex & Events \\ \hline
 & Single & 91 & 2,271 &  901 & 3,759 \\
News & Double & 24 & 570 & 266 & 1,048 \\
& Total & 115 & 2,841 & 1,167 & 4,807\\ \hline
\multirow{3}{*}{Narratives}& Single & 96 & 3,034 &  91 & 9,024 \\
 & Double & 24 & 628 & 40 & 1,952 \\ 
& Total  & 120 & 3,662 &  131 & 10,976 \\ \hline
\end{tabular}
\caption{\label{data-stat} Corpus annotation statistics. ({\em Timex} stands for time expressions.) }
\end{table}

On event annotation, our work is comparable to the annotation work in \newcite{kolomiyets2012extracting}. They report inter-annotator agreements of 0.86, 0.82, and 0.70 on event recognition, unlabeled relations, and labeled relations respectively on a narrative data. We argue that the comparable or better agreements on narratives as shown in Table~\ref{IAA} show that incorporating the notion of linguistic temporal anaphora helps annotators make more consistent decisions. High (above 90\%) agreements on time expression recognition and parsing indicate that our new definition of the reference time for time expressions is clear and easy for annotators to operate on. While event annotations receive lower agreements than time expressions on both genres, they are in general easier on news than on narratives, especially for event reference time resolution and edge labeling.

\begin{table}[h]
\begin{center}
\begin{tabular}{cr|r|r}
\hline
\multicolumn{2}{c|}{ } & News  & Narratives   \\ \hline
\multirow{3}{*}{Timex} & Recognition &  .97 &  1. \\
 & Classification &   .95 &  .94 \\
&  Parsing &  .93 &  .94 \\ \hline 
\multirow{4}{*}{Event} & Recognition &  .94 &  .93 \\
 & Classification &  .77 & .75 \\
& Relations (unlabeled) &  .86 & .83 \\ 
& Relations (labeled) &  .79 &  .72\\ \hline
\end{tabular}
\end{center}
\caption{\label{IAA} Inter-Annotator Agreement F scores on 20\% of the annotations. }
\end{table}

\subsection{Analysis Across Different Genres}
\label{cross-genre}

During our annotation, we discovered that narrative texts are very different from news with respect to their temporal structures. First, news texts are usually organized with abundant temporal locations, while narrative texts tend to start with a few temporal locations setting the scene and proceed with only events. As shown in Table~\ref{data-stat}, around 20\% (1166) nodes in the news data are time expressions and 80\% (4805) are event nodes, while in the narrative data the ratio of time expressions to events are 0.01\%/99.99\% (132/10314). 

Second, descriptive statements are more common in news data than in narratives, while long chains of time advancing eventives are more common in narratives. We can see from Table~\ref{dist-event} that in news data only 30\% events are eventive, leaving the rest 70\% stative descriptions, while in narrative data over half of the events (51\%) are eventive. From Table~\ref{dist-tr} we can also see that the major temporal relation in news is  ``overlap''  (54\%), representing dominative stative statements in reporting discourse mode, while narrative texts are dominated by the ``before'' relation (53\%), with eventive statements advancing the story line. 

\begin{table}[h]
\begin{center}
\begin{tabular}{c|r|r}
\hline
{Timex type} & {News}  & {Narratives}   \\ \hline
Absolute Concrete & 313 (27\%)& 16 (14\%) \\
Relative Concrete &598  (51\%)&  20 (17\%)\\
Vague & 256 (22\%)&  79 (67\%)\\
%Not Locatable Time expressions &  &  \\
\hline
\end{tabular}
\end{center}
\caption{\label{dist-timex} Distribution of time expression types. }
\end{table}

\begin{table}[h]
\begin{center}
\begin{tabular}{c|r|r}
\hline
{Event type} & {News}  & {Narratives }  \\ \hline
Event & 1457 (30\%) & 5594 (51\%) \\
State & 1802 (37\%) &  3366 (31\%)\\
Habitual & 102 (2\%)& 459 (4\%) \\
Modalized & 321 (7\%) & 458 (4\%) \\
Completed & 1041 (22\%) &  900 (8\%) \\
Ongoing Event & 80 (2\%)& 175 (2\%) \\
Generic State & 1 (0\%)& 17 (0\%)\\
Generic Habitual & 2 (0\%)& 5 (0\%)\\\hline
\end{tabular}
\end{center}
\caption{\label{dist-event} Distribution of event types. }
\end{table}

\begin{table}[h]
\begin{center}
\begin{tabular}{c|c|r|r}
\hline
\multicolumn{2}{c|}{Edge label}& {News}  & {Narratives}   \\ \hline
\multicolumn{2}{c|}{Includes} &  1096 (18\%)&  157 (1\%)\\
\multicolumn{2}{c|}{Before(After)} &  507 (8\%) & 5885 (53\%)\\
\multicolumn{2}{c|}{Overlap} &  3246 (54\%) & 4914 (44\%)\\
\multicolumn{2}{c|}{Depend-on} & 1125 (19\%) & 151 (1\%) \\
\hline
%& DCT &  507 (8\%)&  1 ($~$0\%)\\ 
%& Present\_Ref &  493 (8\%)& 11 (0\%)\\ 
%Depend-on & Past\_Ref &  60 (1\%) & 134 (1\%) \\ 
%& Future\_Ref&  26 (0\%)& 3 (0\%)\\ 
%& Atemporal &  39 (1\%)& 2 (0\%)\\ 
%\hline
\end{tabular}
\end{center}
\caption{\label{dist-tr} Distribution of temporal relations. }
\end{table}

Another difference is that statives serve different major roles in news and narrative texts. News tend to have deep branches of overlapping statives with a time expression, DCT, or a general present/past/future reference time as their parent (descriptive statements as discussed above). Narrative texts have much less such long stative branches, however, they tend to have numerous short branches of statives with an eventive event as their parent. These statives serve as the event's accompanying situations. For example, in (\ref{grimm-example}) ``was$_{s4}$'', ``was$_{s5}$'', ``was$_{s6}$'', and ``baptised$_{s7}$'' are accompanying statives to ``came$_{e2}$'', describing the baby and the family and the situation they were in at that time. For each type of node, we compiled the distribution of its possible types of parent, shown in Table~\ref{dist-parent-child}. It's worth noting that more than twice as much statives in news have a stative parent (57\%) than the ones having an eventive parent (24\%), contributing to deep stative branches, while in narratives a much higher percentage of statives directly depend on an eventive (49\%), contributing to a large number of short stative branches.

These different temporal properties of news and narratives further result in shallow dependency structures for news texts with larger number of branches on the root node, yet deep structures for narrative texts with fewer but long branches. These differences are illustrated intuitively on Figure~\ref{full-parse-news} and Figure~\ref{full-parse-grimm}.

%\subsection{Analysis over Event Types}
% TO DO: add analysis over event types

\section{Conclusion}
\label{conclusion}
In this work, we proposed a novel approach to model temporal relations in a document -- building a temporal dependency tree structure for the document. We argue that this structure is linguistically intuitive, and is amenable to computational modeling. High and stable inter-annotator agreements in our annotation experiments provide further evidence supporting our structured approach to temporal interpretation. In addition, a  significant number of documents covering two genres have been annotated. This corpus is publicly available for research on temporal relation analysis, story timeline construction, as well as numerous other applications.

% \nocite{*}
\section{Bibliographical References}
\label{main:ref}

\bibliographystyle{lrec}
\bibliography{xample}

\begin{thebibliography}{}

\bibitem[\protect\citename{Allen}1984]{allen1984towards}
Allen, J.~F.
\newblock (1984).
\newblock Towards a general theory of action and time.
\newblock {\em Artificial intelligence}, 23(2):123--154.

\bibitem[\protect\citename{Bethard \bgroup et al.\egroup }2015]{tempeval2015}
Bethard, S., Derczynski, L., Savova, G., Pustejovsky, J., and Verhagen, M.
\newblock (2015).
\newblock Semeval-2015 task 6: Clinical tempeval.
\newblock In {\em SemEval@ NAACL-HLT}, pages 806--814.

\bibitem[\protect\citename{Bethard \bgroup et al.\egroup }2016]{tempeval2016}
Bethard, S., Savova, G., Chen, W.-T., Derczynski, L., Pustejovsky, J., and
  Verhagen, M.
\newblock (2016).
\newblock Semeval-2016 task 12: Clinical tempeval.
\newblock {\em Proceedings of SemEval}, pages 1052--1062.

\bibitem[\protect\citename{Bethard \bgroup et al.\egroup }2017]{tempeval2017}
Bethard, S., Savova, G., Palmer, M., and Pustejovsky, J.
\newblock (2017).
\newblock Semeval-2017 task 12: Clinical tempeval.
\newblock In {\em Proceedings of the 11th International Workshop on Semantic
  Evaluation (SemEval-2017)}, pages 565--572, Vancouver, Canada, August.
  Association for Computational Linguistics.

\bibitem[\protect\citename{Bohnemeyer}2009]{bohnemeyer2009temporal}
Bohnemeyer, J.
\newblock (2009).
\newblock Temporal anaphora in a tenseless language.
\newblock {\em The expression of time in language}, pages 83--128.

\bibitem[\protect\citename{Chambers and Jurafsky}2008]{chambers2008jointly}
Chambers, N. and Jurafsky, D.
\newblock (2008).
\newblock Jointly combining implicit constraints improves temporal ordering.
\newblock In {\em Proceedings of the Conference on Empirical Methods in Natural
  Language Processing}, pages 698--706. Association for Computational
  Linguistics.

\bibitem[\protect\citename{Do \bgroup et al.\egroup }2012]{do2012joint}
Do, Q.~X., Lu, W., and Roth, D.
\newblock (2012).
\newblock Joint inference for event timeline construction.
\newblock In {\em Proceedings of the 2012 Joint Conference on Empirical Methods
  in Natural Language Processing and Computational Natural Language Learning},
  pages 677--687. Association for Computational Linguistics.

\bibitem[\protect\citename{Hinrichs}1986]{hinrichs1986temporal}
Hinrichs, E.
\newblock (1986).
\newblock Temporal anaphora in discourses of english.
\newblock {\em Linguistics and philosophy}, 9(1):63--82.

\bibitem[\protect\citename{Kolomiyets \bgroup et al.\egroup
  }2012]{kolomiyets2012extracting}
Kolomiyets, O., Bethard, S., and Moens, M.-F.
\newblock (2012).
\newblock Extracting narrative timelines as temporal dependency structures.
\newblock In {\em Proceedings of the 50th Annual Meeting of the Association for
  Computational Linguistics: Long Papers-Volume 1}, pages 88--97. Association
  for Computational Linguistics.

\bibitem[\protect\citename{Mostafazadeh \bgroup et al.\egroup
  }2016]{mostafazadeh2016caters}
Mostafazadeh, N., Grealish, A., Chambers, N., Allen, J., and Vanderwende, L.
\newblock (2016).
\newblock Caters: Causal and temporal relation scheme for semantic annotation
  of event structures.
\newblock In {\em Proceedings of the The 4th Workshop on EVENTS: Definition,
  Detection, Coreference, and Representation, San Diego, California, June.
  Association for Computational Linguistics}.

\bibitem[\protect\citename{O'Gorman \bgroup et al.\egroup }2016]{o2016richer}
O'Gorman, T., Wright-Bettner, K., and Palmer, M.
\newblock (2016).
\newblock Richer event description: Integrating event coreference with
  temporal, causal and bridging annotation.
\newblock {\em Computing News Storylines}, page~47.

\bibitem[\protect\citename{Partee}1973]{partee1973some}
Partee, B.~H.
\newblock (1973).
\newblock Some structural analogies between tenses and pronouns in english.
\newblock {\em The Journal of Philosophy}, 70(18):601--609.

\bibitem[\protect\citename{Partes}1984]{partes1984nominal}
Partes, B.~H.
\newblock (1984).
\newblock Nominal and temporal anaphora.
\newblock {\em Linguistics and philosophy}, 7(3):243--286.

\bibitem[\protect\citename{Pustejovsky \bgroup et al.\egroup
  }2003a]{pustejovsky2003timeml}
Pustejovsky, J., Castano, J.~M., Ingria, R., Sauri, R., Gaizauskas, R.~J.,
  Setzer, A., Katz, G., and Radev, D.~R.
\newblock (2003a).
\newblock Timeml: Robust specification of event and temporal expressions in
  text.
\newblock {\em New directions in question answering}, 3:28--34.

\bibitem[\protect\citename{Pustejovsky \bgroup et al.\egroup
  }2003b]{pustejovsky2003timebank}
Pustejovsky, J., Hanks, P., Sauri, R., See, A., Gaizauskas, R., Setzer, A.,
  Radev, D., Sundheim, B., Day, D., Ferro, L., et~al.
\newblock (2003b).
\newblock The timebank corpus.
\newblock In {\em Corpus linguistics}, volume 2003, page~40. Lancaster, UK.

\bibitem[\protect\citename{Reichenbach}1947]{reichenbach47}
Reichenbach, H.
\newblock (1947).
\newblock {\em {Elements of Symbolic Logic}}.
\newblock The MacMillan Company, New York.

\bibitem[\protect\citename{Smith}2003]{smith2003modes}
Smith, C.~S.
\newblock (2003).
\newblock {\em Modes of discourse: The local structure of texts}, volume 103.
\newblock Cambridge University Press.

\bibitem[\protect\citename{Styler~IV \bgroup et al.\egroup
  }2014]{styler2014temporal}
Styler~IV, W.~F., Bethard, S., Finan, S., Palmer, M., Pradhan, S., de~Groen,
  P.~C., Erickson, B., Miller, T., Lin, C., Savova, G., et~al.
\newblock (2014).
\newblock Temporal annotation in the clinical domain.
\newblock {\em Transactions of the Association for Computational Linguistics},
  2:143--154.

\bibitem[\protect\citename{UzZaman \bgroup et al.\egroup }2012]{tempeval2012}
UzZaman, N., Llorens, H., Allen, J., Derczynski, L., Verhagen, M., and
  Pustejovsky, J.
\newblock (2012).
\newblock Tempeval-3: Evaluating events, time expressions, and temporal
  relations.
\newblock {\em arXiv preprint arXiv:1206.5333}.

\bibitem[\protect\citename{Verhagen \bgroup et al.\egroup }2007a]{tempeval2007}
Verhagen, M., Gaizauskas, R., Schilder, F., Hepple, M., Katz, G., and
  Pustejovsky, J.
\newblock (2007a).
\newblock Semeval-2007 task 15: Tempeval temporal relation identification.
\newblock In {\em {Proceedings of the 4th International Workshop on Semantic
  Evaluations}}, pages 75--80. Association for Computational Linguistics.

\bibitem[\protect\citename{Verhagen \bgroup et al.\egroup
  }2007b]{verhagen2007semeval}
Verhagen, M., Gaizauskas, R., Schilder, F., Hepple, M., Katz, G., and
  Pustejovsky, J.
\newblock (2007b).
\newblock Semeval-2007 task 15: Tempeval temporal relation identification.
\newblock In {\em Proceedings of the Fourth International Workshop on Semantic
  Evaluations (SemEval-2007)}, pages 75--80, Prague, Czech Republic, June.
  Association for Computational Linguistics.

\bibitem[\protect\citename{Verhagen \bgroup et al.\egroup }2009]{verhagen09}
Verhagen, M., Gaizauskas, R., Schilder, F., Hepple, M., Moszkowicz, J., and
  Pustejovsky, J.
\newblock (2009).
\newblock {The TempEval Challeage: Identifying Temporal Relations in Text}.
\newblock {\em Lang Resources \& Evaluation}, 43:161--179.

\bibitem[\protect\citename{Verhagen \bgroup et al.\egroup }2010a]{tempeval2010}
Verhagen, M., Sauri, R., Caselli, T., and Pustejovsky, J.
\newblock (2010a).
\newblock Semeval-2010 task 13: Tempeval-2.
\newblock In {\em Proceedings of the 5th international workshop on semantic
  evaluation}, pages 57--62. Association for Computational Linguistics.

\bibitem[\protect\citename{Verhagen \bgroup et al.\egroup
  }2010b]{verhagen2010semeval}
Verhagen, M., Sauri, R., Caselli, T., and Pustejovsky, J.
\newblock (2010b).
\newblock Semeval-2010 task 13: Tempeval-2.
\newblock In {\em Proceedings of the 5th International Workshop on Semantic
  Evaluation}, pages 57--62, Uppsala, Sweden, July. Association for
  Computational Linguistics.

\bibitem[\protect\citename{Webber}1988]{webber1988tense}
Webber, B.~L.
\newblock (1988).
\newblock Tense as discourse anaphor.
\newblock {\em Computational Linguistics}, 14(2):61--73.

\bibitem[\protect\citename{Wuyun}2016]{Wuyun201638}
Wuyun, S.
\newblock (2016).
\newblock The influence of tense interpretation on discourse coherence - a
  comparison between mandarin narrative and report discourse.
\newblock {\em Lingua}, 179:38 -- 56.

\bibitem[\protect\citename{Xue \bgroup et al.\egroup }2005]{xue2005nle}
Xue, N., Xia, F., Chiou, F.-D., and Palmer, M.
\newblock (2005).
\newblock The penn chinese treebank: Phrase structure annotation of a large
  corpus.
\newblock {\em Natural language engineering}, 11(2):207--238.

\bibitem[\protect\citename{Xue \bgroup et al.\egroup }2016]{xue2016conll}
Xue, N., Ng, H.~T., Pradhan, S., Rutherford, A., Webber, B., Wang, C., and
  Wang, H.
\newblock (2016).
\newblock Conll 2016 shared task on multilingual shallow discourse parsing.
\newblock {\em Proceedings of the CoNLL-16 shared task}, pages 1--19.

\bibitem[\protect\citename{Yoshikawa \bgroup et al.\egroup
  }2009]{yoshikawa2009jointly}
Yoshikawa, K., Riedel, S., Asahara, M., and Matsumoto, Y.
\newblock (2009).
\newblock Jointly identifying temporal relations with markov logic.
\newblock In {\em Proceedings of the Joint Conference of the 47th Annual
  Meeting of the ACL and the 4th International Joint Conference on Natural
  Language Processing of the AFNLP: Volume 1-Volume 1}, pages 405--413.
  Association for Computational Linguistics.

\bibitem[\protect\citename{Zhang and Xue}2014]{zhang2014automatic}
Zhang, Y. and Xue, N.
\newblock (2014).
\newblock Automatic inference of the tense of chinese events using implicit
  linguistic information.
\newblock In {\em EMNLP}, pages 1902--1911. Citeseer.

\end{thebibliography}

%\section{Language Resource References}
%\label{lr:ref}
%\bibliographystylelanguageresource{lrec}
%\bibliographylanguageresource{xample}

\end{document}